\pdfoutput=1

\documentclass[11pt]{article}

\usepackage[preprint]{acl}

\usepackage{times}
\usepackage{latexsym}

\usepackage[T1]{fontenc}

\usepackage[utf8]{inputenc}

\usepackage{microtype}

\usepackage{inconsolata}

\usepackage{graphicx}

\title{Enhancing Persona Consistency for LLMs' Role-Playing using Persona-Aware Contrastive Learning}

\author{Ke Ji$^{1,2}$\thanks{Work done during an internship at Xiaobing.AI}, Yixin Lian$^{2}$, Linxu Li$^{2}$, Jingsheng Gao$^{2,3}$, Weiyuan Li$^{2}$, \textbf{Bin Dai}$^{2}$\thanks{Corresponding Author} \\
$^1$ Southeast University, $^2$ Xiaobing.AI, $^3$ Shanghai Jiao Tong University \\
}

\usepackage{xcolor}
\usepackage{tcolorbox}
\usepackage{booktabs}
\usepackage{multicol}
\usepackage{multirow}
\usepackage{array}
\usepackage{longtable}
\usepackage{amsmath}
\usepackage{amssymb}
\usepackage{pifont}
\usepackage{graphicx}

\newcommand{\cmark}{\ding{51}} 
\newcommand{\xmark}{\ding{55}} 

\newtcolorbox{alprompt}[1]{
        boxrule = 1pt,
        fontupper = \small\tt,
        fonttitle = \bf\color{black},
        arc = 2pt,
        rounded corners,
        colframe = black,
        colbacktitle = white!97!yellow,
        colback = white!97!yellow,
        title = #1,
}

\definecolor{outerboxcolor}{gray}{0.90} 
\definecolor{innerboxcolor}{rgb}{1,1,1}

\begin{document}
\maketitle
\begin{abstract}
In recent years, large language models (LLMs) have achieved breakthrough progress in many dialogue generation tasks. However, their lack of emotion and fine-grained role awareness limits the model's ability to provide personalized and diverse interactions further. Current methods face high costs in collecting high-quality annotated data for scenarios such as role-playing, and traditional human alignment methods are difficult to deploy due to the inherent diversity of model behavior in role-playing scenarios.
Inspired by the alignment of models for safety behaviors through RLHF (Reinforcement Learning from Human Feedback), in this paper, we revisit model role-playing behavior from the perspective of persona alignment and propose a novel annotation-free framework named \textbf{\underline{P}}ersona-Aware \textbf{\underline{C}}ontrastive \textbf{\underline{L}}earning (PCL) to align LLMs' behavior during role-playing, enhancing the model's role consistency. Specifically, we first design a role chain method to encourage the model to self-question based on the role characteristics and dialogue context to adjust personality consistency. 
Then, we further enhance the model's role-playing strategy through iterative contrastive learning between the use of role characteristics and not.
Experiments on both black-box and white-box LLMs show that LLMs equipped with PCL significantly outperform vanilla LLMs under automatic evaluation methods (CharEval \& GPT-4) and human expert evaluation.
\end{abstract}

\section{Introduction}

Due to the success of self-supervised learning and instruction tuning, large language models (LLMs) have achieved remarkable success in many NLP tasks, 
such as text generation~\cite{iqbal2022survey}, instruction following~\cite{wang2023aligning}, summarization~\cite{tam-etal-2023-evaluating}, and other applications~\cite{yang2020survey,ji-etal-2023-hierarchical}.
With the growing demand for diversity in psychology and entertainment, more and more research is focusing on how to guide LLMs to cultivate personalized imitation and role-playing capabilities, rather than merely providing knowledge interaction.
However, an AI assistant designed to accomplish various knowledge interaction tasks finds it difficult to provide satisfactory personalized interactions due to the lack of emotions and perception of the real world, limiting the human imagination of applying LLMs in more diverse scenarios.

Compared with the utilization of language models as general assistants for answering broad knowledge questions, the main difference of role-playing lies in that the key point is to keep the LLM in line with the given specific role profiles~\cite{tu2024charactereval, tu2023characterchat, chen2023large}.
To incorporate the character's personality into the model interaction, 
early studies often adopted customized designs~\cite{zhang-etal-2018-personalizing,jiang-etal-2024-personallm,DBLP:conf/ijcai/KotturWC17,DBLP:conf/acl/LiGBSGD16}, such as using memory networks or custom transformers to effectively fuse storage- and embedding-based personal information.
Nevertheless, because of their highly specialized designs, such studies are deficient in versatility and scalability.

Benefiting from LLM's powerful ability to understand context and generate corresponding responses, the dialogue can remain coherent and context-rich even with minimal prompts.
\citet{chen2023large} and \citet{tu2024charactereval} customized various roles by configuring personal background profiles within their prompts, intending to closely emulate specific characters.
When the virtual world where the character needs to be played is very different from the real world,
traditional pre-training corpora struggle to effectively incorporate the knowledge required for role-playing, thus leading to character hallucination~\cite{shao2023character,ahn2024timechara,sadeq2024mitigating,gao2023livechat}.
Inspired by the domain adaptation methods~\cite{kenton2019bert,Radford2019LanguageMA,brown2020language,touvron2023llama}, which enhances the model's understanding of domain knowledge by collecting a large number of unlabeled predictions and using large-scale self-supervised pre-training, \citet{Zheng_Zhang_Huang_Mao_2020}, \citet{ wang2023rolellm} and \citet{zhou2023characterglm} incorporated the necessary background knowledge for role profile through pre-training.
However, due to the disparity in basic common sense, knowledge, and viewpoints between the backbone model trained based on the real-world corpus and the character we want to imitate
, such a pre-training process may undermine the generalization potential of the backbone model and further constrain the comprehensive performance of LLM in role-playing.

Furthermore, CharacterGLM \cite{zhou2023characterglm} and Ditto \cite{lu2024large} developed character-specific LLMs based on instruction-tuning technology to enhance the model's fine-grained perception of role-playing roles.
\citet{shea-yu-2023-building} employed offline reinforcement learning strategies named RLHF~\cite{DBLP:conf/nips/Ouyang0JAWMZASR22} to enhance persona consistency.
Similarly, COMEDY \cite{chen2024compress} also used GPT-4 to generate a large amount of character preference data and deployed DPO~\cite{rafailov2024direct} method to align models to produce more coherent memory-based personalized interactions.
However, collecting and annotating high-quality preference data is more sophisticated than traditional tasks, as it requires annotators to have a profound understanding of specific characters to precisely label preferences~\cite{chen2024oscars}. 
Moreover, the variations in the interpretation of characters among annotators can cause inconsistencies, making it hard to scale such methods effectively.

Despite the significant amount of research conducted on role-playing based on LLMs, current studies still face several major challenges:
1) The collection of large-scale, high-quality role-playing datasets is costly, making existing methods that rely on massive manual annotations difficult to scale effectively.
2) The supervisory signals in role enactment scenarios are usually implicit and diverse, making it challenging to directly and effectively align the model's role enactment behavior.
3) Besides, it is worth noting that extensive role-specific fine-tuning might impair the model's generalization potential and weaken its commonsense reasoning capabilities. This, in turn, further constrains the model's role-playing abilities.

In this paper, we hence propose a novel annotation-free framework named 
Persona-Aware Contrastive Learning (PCL)
to align LLMs' behavior during role enactment scenario.
Specifically, we initially design a chain of persona self-reflections to encourage the model to conduct self-questioning based on dialogue context and the role's characteristics to align with personality consistency. 
Then we improve the model's persona-driven strategy by progressively adopting the contrasitve learning process, alternating between the application of role characteristics and their absence.

To sum up, the main contributions of this paper are listed as follows:
\begin{itemize}
    \item We introduce an innovative annotation-free framework that can align the model's role-based behavior, thereby improving persona consistency at a lower cost.
    \item The proposed framework re-examines the problem of behavioral supervision of role-playing from the perspective of AI style alignment, and gradually enhances the role-playing strategy of the model through self-play contrastive training.
    
    \item We perform PCL on both black- and white-box LLMs and extensive experiments show that LLMs equipped with PCL significantly outperform vanilla LLMs using GPT-4 and CharEval evaluations while maintaining comparable levels of general knowledge.
\end{itemize}

\section{Preliminaries}
The Role-Playing Conversational Agent (RPCA) is designed to engage in conversations with users by emulating specific characters. These characters are defined by their knowledge, behavior, and style of response. To achieve this, the RPCA utilizes a character profile, denoted as $P$, and the current dialogue context, represented as $C_n = [c_1, c_2, \dots, c_n]$. 
Here, $c_i$ corresponds to the $i$-th utterance in the dialogue, respectively. The goal for the RPCA is to generate a response $y$ that is consistent with the character's profile, which can be represented as:
\begin{equation}
y = \texttt{RPCA}(C_n, P),
\end{equation}
In this paper, we define role-playing tasks by providing large models with specific character names or brief descriptions. 
Subsequently, we assess their ability to maintain consistent self-awareness and demonstrate nuanced character-specific knowledge across multiple rounds of dialogue.

\section{Related Work}
\subsection{Human Alignment}
To align the LLMs with human preference of honesty,  helpfulness and harmlessness, etc, various methods have been proposed to achieve this goal, such as reinforcement learning from human feedback (RLHF)~\cite{bai2022training}. Later works ~\cite{rafailov2024direct, azar2023generaltheoreticalparadigmunderstand, ahmadian2024basicsrevisitingreinforcestyle} have further proposed optimization methods on this basis. Nevertheless, these methods heavily rely on the human-annotated preference. Although several works ~\cite{munos2024nashlearninghumanfeedback, calandriello2024humanalignmentlargelanguage} have proposed leveraging online data as an alternative to regularisation, they still require supervised preference data at the beginning of training. To alleviate the requirements of labeled preferences, ~\citet{lee2023rlaif} proposed RLAIF to use off-the-shelf LLMs to substitute humans in labeling preferences. 
However, the noise of the resulting labeled preferences appears unstable, and the robustness is not guaranteed when selecting different off-the-shelf LLMs.
It should be noted that \citet{burns2023weak} has suggested that the human alignment of future superhuman models may become challenging because of the models' complex behaviors, leading to the ineffectiveness of traditional alignment methods. This viewpoint is highly applicable to role-based scenarios, where human annotation is extremely difficult and costly. 

\subsection{Chain-of-Thought}
Chain-of-Thought (CoT), first proposed by \citet{wei2023chainofthoughtpromptingelicitsreasoning}, has strikingly enhanced the reasoning abilities of LLMs. The method emphasizes enriching responses with detailed reasoning steps, thus conspicuously improving the performance on reasoning tasks. Later works, have further refined the CoT approach, such as Self-Ask~\cite{press-etal-2023-measuring}, Self-Consistency ~\cite{wang2023selfconsistencyimproveschainthought}, Least-to-Most prompting~\cite{zhou2023leasttomostpromptingenablescomplex}, and Tree of Thought~\cite{NEURIPS2023_271db992}. Recently, some works have begun to introduce CoT in dialogue problems. For example, Cue-CoT~\cite{wang-etal-2023-cue} proposed to explore the underlying linguistic cues about the user status exhibited in the context. DOCTOR~\cite{chae-etal-2023-dialogue} claimed that CoT prompting in dialogues is a non-trivial challenge and put forward multi-hop reasoning in dialogue. In this paper, we further investigate the advantage of appropriate CoT prompting in persona dialogue and demonstrate its effectiveness in enhancing role consistency.

\subsection{Role-Playing}
Recent advancements in the LLM community have highlighted the potential of LLM customization and role-playing \cite{wei2023multiparty, shanahan2023roleplay, li2023chatharuhi, salemi2023lamp, fable2023showrunner, camel, chen2023autoagent, generative-agents}. 
Adopting specific roles allows LLMs to exhibit enhanced vividness \cite{li2023chatharuhi}, interactivity \cite{fable2023showrunner, wang2023interactive}, personalization \cite{salemi2023lamp}, and capability in tackling complex tasks \cite{camel, chen2023autoagent, qian2023communicative}. 
Nevertheless, open-source LLMs still significantly trail behind state-of-the-art closed-source counterparts like GPT-4 in terms of role-playing proficiency.

\begin{figure*}[t]
\centering
\includegraphics[width=1.0\textwidth]{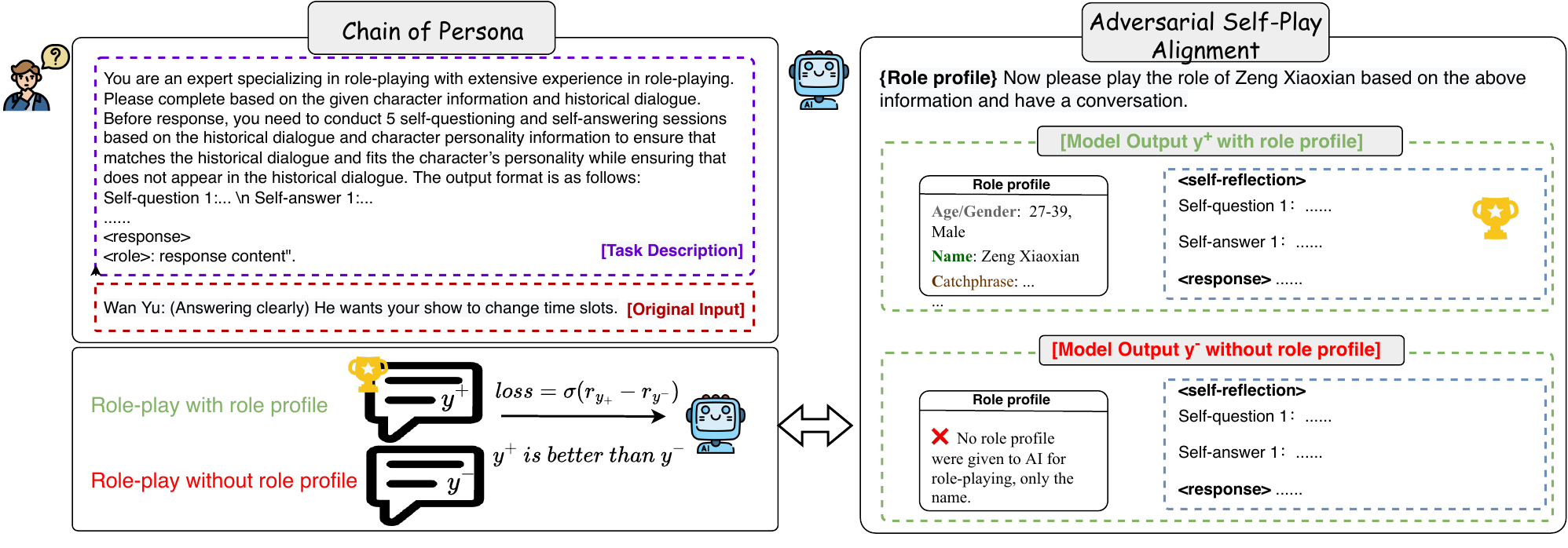}
\caption{
Our proposed PCL adopts chain of persona, and generates contrastive samples by removing the necessary role profile for self-play alignment. This self-play process is conducted iteratively through different epochs. 
}
\label{fig:AL_Overview}
\end{figure*}

\section{Method}

\subsection{Overview}
\label{sec:overview}

In this section, we give an overview of the proposed \textsc{Persona-Aware Contrastive Learning} (PCL).
Initially, we design a chain of persona to prompt the model to engage in self-questioning through dialogue, ensuring persona consistency, as shown in Section~\ref{sec:cop}.
Subsequently, we have the model output self-questioning twice, one gives the response without relying on the given persona, and the other is based on the given persona, and then engages the model in contrastive learning based on these two role-plays to align the model's consistent persona behavior during role-playing, as shown in Section~\ref{sec:asp}.

\begin{figure}[t]
\centering
\scriptsize
\begin{alprompt}{\centering \footnotesize \textit{The prompt for chain of persona}}
You are an expert specializing in role-playing with extensive experience in role-playing. 
Please complete <response> based on the given character information and historical dialogue. 
Before response, you need to conduct 5 self-questioning and self-answering sessions based on the historical dialogue and character personality information to ensure that <response> matches the historical dialogue and fits the character's personality while ensuring that <response> does not appear in the historical dialogue. The output format is as follows: \newline
"<self-questioning and self-answering> \newline
Self-question 1:...\newline
Self-answer 1:...\newline
...\newline
Self-question 5:...\newline
Self-answer 5:...\newline
<response>\newline
<role name>: response content".\newline
\newline
\{\colorbox{outerboxcolor}{\color{purple} [role profile]} \} \newline
Now please play the role of \colorbox{outerboxcolor}{\color{purple} [role name]} based on the above role profile and have a conversation.\newline
\colorbox{outerboxcolor}{\color{teal} [dialogue history]}
\end{alprompt}
\caption{\label{fig:cop_prompt} The prompt for generating self-reflection persona constraints and the final response.}
\end{figure}

\subsection{Chain of Persona Design}
\label{sec:cop}
An LLM takes a token sequence $x$ as input and defines a conditional distribution over an output sequence $y$. The conditional distribution can be represented as $\pi_\theta(y|x)$, with $\theta$ being the parameter of the LLM.

Generally, given a dialogue history $C_n$ and character profile $P$, we have
\begin{equation}
    y \sim \pi_\theta\left(\cdot|\text{Prompt}(P, C_n)\right)
\end{equation}

Inspired by COT, chain of persona (COP) is a special design of the prompt such that it can constrain the model's responses to ensure they align with the persona of the character being played. 
We achieve this goal by having the model complete a series of persona-aware self-reflections before generating a response. 
Therefore, the output of COP is considered to comprise two parts: \textit{self-reflection persona constraints} and \textit{final role-playing response}.

The detailed design of COP prompt is shown in Figure~\ref{fig:cop_prompt}, for other possible COP template, we study the impact of template design to evaluate its robustness in Appendix~\ref{appendix:template}. 
Suppose the model can well follow the instruction of our COP prompt, any sampled output $y$ can be decomposed into the following format
\begin{equation}
    y = [q_1, a_1, \dots, q_t, a_t, a_f]
    \label{eqn:format}
\end{equation}
where $q_t$ denote the $i$-th self-questioning, $a_t$ is the answer corresponding to the $i$-th question and $a_f$ represents the final role-playing response.

To facilitate understanding, $\{q_t,a_t\}$ can be seen as the intermediate thought process in the model before making the final role-playing response.
Meanwhile, $t$ denotes the number of self-reflections for persona-aware self-questioning and self-answering.

\subsection{Persona-Aware Training}
\label{sec:self_cop}
Unfortunately, less capable LLMs like QWen-7B and Baichuan2-7B cannot always follow our COP prompt, making COP prompt fail to generate the answer with the correct format. In this scenario, we can apply persona-aware training with the help of more capable LLMs like GPT. 

Denote the GPT model as $\pi_{GPT}$. We first collect 1000 data points $(P, C_n)$ and produce the COP prompt $\text{COP}(P, C_n)$ for each sample. Then we randomly sample an output $y$ from the distribution $\pi_{GPT}(\cdot|\text{COP}(P, C_n))$ for each data point, such that $y$ follows the format in (\ref{eqn:format}). 
We conducted experiments on the impact of warmup data in the Appendix~\ref{appendix:warmup}.
The persona-aware training objective then becomes
\begin{equation}
    \mathcal{L}(P, C_n) =  -\log \pi_\theta(y| \text{COP}(P, C_n))
\end{equation}
which is similar to the vanilla supervised fine-tuning (SFT) objective.
An example of our persona-aware training data is shown in Table~\ref{tab:SFT_example}. 

\subsection{Contrastive Self-Play Alignment}
\label{sec:asp}
COP can be seen as a skill for role-playing in models, which helps the model better align character attributes and conversation topics.
To further supervise the model's behavior during role-playing, inspired by human contrastive self-reflection to strengthen their understanding of roles and thereby enhance their role-playing abilities, we have designed an contrastive self-play alignment (ASPA) method that does not require additional preference annotations.

Specially, 
given a dialogue history $C_n$ and a role profile $P$, 
we can sample a preference pair $(y^+, y^-)$ using the following method
\begin{align}
    y^+ & \sim \pi_\theta\left(\cdot|\text{COP}(P, C_n)\right) \\
    y^- & \sim \pi_\theta\left(\cdot|\text{COP}(P=\emptyset, C_n)\right)
\end{align}
where $P=\emptyset$ means the character profile is empty.
With this preference pair, we use a Direct Preference Optimization (DPO)~\cite{rafailov2024direct} strategy to help the LLM perform contrastive self-play alignment.
The optimization objective is as follows:

\begin{align}
\label{eq:ASRA_obj}
& \mathcal{L}_\text{ASPA}(P, C_n) =  \nonumber - \log \sigma \Bigg( \beta \log \frac{\pi_{\theta}(y^+ \mid x)}{\pi_{\theta_{ref}}(y^+ \mid x)} \\
& \quad - \beta \log \frac{\pi_{\theta}(y^- \mid x)}{\pi_{\theta_{ref}}(y^- \mid x)} \Bigg)
\end{align}
where $\theta_{ref}$ is the reference model parameter, $x$ is the COP prompt $\text{COP}(P, C_n)$ while $\beta$ is a hyperparameter for DPO.

After each epoch, the model generates \{y\} for all given $\{P,C_n\}$ for the next epoch training, so that it can be iteratively optimized through self-play and no additional external annotation is required.

\section{Experiments}

\subsection{Experiments Setup}

\subsubsection{Dataset and Experiments Setting}
\label{sec:exper_setting}
We conduct extensive experiments using a more challenging CharacterEval~\cite{tu2024charactereval} benchmark.
CharacterEval is a role-playing dataset based on multi-turn dialogues, in which models need to perform role-playing based on the given persona information and dialogue history.

To thoroughly emphasize the proposed PCL, we consider two settings to measure the proposed methods: general setting and transfer setting.
In our experiments, PCL is first trained on the training set of the dataset and then evaluated on the test set.
\paragraph{General setting}
We follow the original setup proposed in CharacterEval where the whole 77 character profiles appear both in the train and test sets.
\paragraph{Transfer setting}
We aim to investigate the role-playing generalization of the proposed method.
Specifically, we split the original dataset into two separate sets, and these two sets contain non-overlapping sets of character profiles.
We randomly choose 60 character profiles and their corresponding dialogues as the training set, and select the remaining 17 character profiles and corresponding dialogues as the test set.

When deploying PCL on open source models with fewer parameters, we first perform persona-aware training and then perform contrastive self-play alignment.
After that we use COP as shown in Figure~\ref{fig:cop_prompt} to prompt the model on test set.
Note that there is no need of trainset annotation for PCL.

\subsubsection{LLM baselines}

\begin{table}[t]
\centering
\small
\renewcommand{\arraystretch}{0.8}  
\setlength{\tabcolsep}{1.1mm}
\begin{tabular}{l|cccc}
\toprule
\textbf{Models} & \textbf{Specialized} & \textbf{IsOpen} &  \textbf{Main Language} \\
\midrule
Qwen-7B  & \xmark & \cmark & zh \\
Baichuan2-7B & \xmark & \cmark & zh\\
CharacterGLM-6B & \cmark  & \cmark  & zh  \\
GPT-3.5 & \xmark &  \xmark& en \\
GPT-4 & \xmark & \xmark& en \\
\bottomrule
\end{tabular}
\caption{LLMs evaluated in our experiments.}
\label{tab:llms}
\end{table}
Experiments mainly focus on advanced LLMs like GPT-3.5~\cite{chatgpt} (GPT-3.5-turbo-1106), and GPT-4~\cite{openai2023gpt4} (GPT-4-0125-preview), as well as open-source models including 
Qwen-7B~\cite{bai2023qwen}, Baichuan2-7B~\cite{yang2023baichuan}, and CharacterGLM-6B~\cite{zhou2023characterglm}.
CharacterGLM-6B is designed specifically for role playing.
Table~\ref{tab:llms} shows the details of used LLMs.
For Qwen-7B and Baichuan2-7B, we use \texttt{Qwen/Qwen-7B-Chat} and \texttt{baichuan-inc/Baichuan2-7B} on hugginceface\footnote{https://huggingface.co/}.
For closed-source models, we directly deploy Chain of Persona.

We consider incorporating the following methods, which are widely used in role-playing scenarios, into our experiments:
\textbf{In-context learning (ICL)}~\cite{wei2022chain}: Following previous work~\cite{tu2024charactereval}, we directly use in-context learning (ICL) to perform role-playing for baseline models, the prompt used for in-context learning is shown in Figure~\ref{fig:in_context_learning} and Table~\ref{tab:in_context_sft}.
\textbf{Vanilla Supervised fine-tuning (SFT)}~\cite{ouyang2022training}: We also apply the supervised fine-tuning (also called as instruction tuning in some cases) method which is widely used by many of the recent role-playing studies to inject role information, such as RoleLLM~\cite{wang2023rolellm}, CharacterGLM~\cite{zhou2023characterglm}, RoleCraft-GLM~\cite{tao2024rolecraft}.

To implement SFT method in our study, we use templates as shown in Table~\ref{tab:in_context_sft} to wrap the training set to generate our SFT dataset, selecting the last round of dialogue as the model’s role-playing response, and then train the model based on the vanilla SFT objective.

Existing research primarily focuses on how to collect role-playing based dialogue corpora for instruction tuning through various heuristic strategies and extensive efforts. 
Therefore, it is worth noting that our approach (PCL) is orthogonal to existing role-playing methods, rather than replacing them. 
PCL can further enhance the role-playing capabilities of existing models and methods.

\subsection{Evaluation metrics}
To conduct a more comprehensive evaluation to measure the role-play ability of the Role-Playing Conversational Agents (RPCA), we adopt the evaluation metrics conducted by~\citet{tu2024charactereval}:
\paragraph{Conversational Ability (CA):}
This part focuses on three key objectives for generated responses: fluency, coherency, and consistency~\cite{zhang2021dynaeval, mesgar2020improving}.
\begin{itemize}
    \item \textbf{Fluency (Flu.)} measures the grammatical correctness and readability of a response. 
    \item \textbf{Coherency (Coh.)} evaluates the relevance of a response to the given context or topic. 
    \item \textbf{Consistency (Cons.)} assesses whether responses remain stable and non-contradictory throughout a conversation.
\end{itemize}    
\paragraph{Character Consistency (CC):}
Character consistency is essential for evaluating the role-play ability of RPCAs, significantly affecting the user experience. 
We assess this via the following metrics:
\begin{itemize}
    \item \textbf{Knowledge-Exposure (KE)}:
    An RPCA should reflect relevant knowledge in its responses to support further evaluation of its knowledge expression capabilities. 
    
    \item \textbf{Knowledge-Accuracy (KA)}: After demonstrating knowledge, an RPCA must generate responses that accurately align with the character's profile. 
    \item \textbf{Knowledge-Hallucination (KH)}: 
    Inspired by studies on LLM hallucinations, an RPCA should avoid incorporating unknown or incorrect knowledge in its responses. 
    \item \textbf{Persona-Behavior (PB)}: Consistent behaviors, described within brackets, improve the user's experience by accurately portraying the character's actions, expressions, and tones.
    \item \textbf{Persona-Utterance (PU)}: 
    An RPCA's utterances should align with the character's habitual expressions to mimic the character.
\end{itemize}

\paragraph{Role-playing Attractiveness (RA):}
As an entertainment conversational agent, an RPCA must be sensitive to user emotions. 
We introduce the following metrics:
\begin{itemize}
    \item \textbf{Human-Likeness (HL)} requires RPCAs to avoid robotic and emotionless responses.
    \item \textbf{Communication Skills (CS)} involve demonstrating a high Emotional Quotient (EQ), making RPCAs more engaging and likable.
    \item \textbf{Expression Diversity (ED)} necessitates RPCAs to display diverse expressive abilities, to provide a more immersive user experience.
    \item \textbf{Empathy (Emp.)} involves expressing empathy, making RPCAs appear warmer.
\end{itemize}

\begin{table}[b]
\renewcommand{\arraystretch}{1}  
\centering
\small
\begin{tabular}{l|cc}
\toprule
Hyperparameters & COP &CSPA  \\
\midrule
Optimizer  & adamw &rmsprop   \\
Warmup Ratio  & 0.1 & 0.2   \\
Learning Rate  & 5e-6 &5e-7   \\
LR Schedule & cosine &linear \\
Batch Size & \multicolumn{2}{c}{8}        \\
Max Length & \multicolumn{2}{c}{2048}     \\
Epoch  & 3 &2    \\
\bottomrule
\end{tabular}
\caption{The hyperparameters of our PCL.}
\label{tab:hyper_PCL}
\end{table}

\begin{table*}[t!]
\renewcommand{\arraystretch}{0.8}  
\centering
\small
\begin{tabular}{l|>{\centering\arraybackslash}m{0.91cm}>{\centering\arraybackslash}m{0.91cm}>{\centering\arraybackslash}m{0.91cm}>{\centering\arraybackslash}m{0.91cm}>{\centering\arraybackslash}m{0.91cm}>{\centering\arraybackslash}m{0.91cm}>{\centering\arraybackslash}m{1.8cm}}

\toprule[1pt]

\multicolumn{1}{c|}{\multirow{2}{*}{Model}}  & \multicolumn{6}{c}{\textbf{Character Consistency}} \\ 
\cmidrule{2-7}
& \multicolumn{1}{c}{KE} & \multicolumn{1}{c}{KA} & \multicolumn{1}{c}{KH} & \multicolumn{1}{c}{PB} & \multicolumn{1}{c}{PU} & \multicolumn{1}{c}{Avg.} \\
\midrule
Baichuan2-7B+ICL & 1.809 & 2.853 & 2.919 & {2.840}  & \textbf{3.081} & 2.700 \\
Baichuan2-7B+SFT+ICL &2.115 & 3.067 & 3.050 &2.511 &2.967 &2.742 \\
Baichuan2-7B+PCL (Ours) &2.268  &\textbf{3.125}  &\textbf{3.131} & 2.454 &3.018 & \textbf{2.799}\\
\midrule
Qwen-7B+ICL & 1.956 & 2.728 & 2.633 & 2.605 & 2.780 & 2.540  \\

Qwen-7B+SFT+ICL &2.336 & 2.945 & 2.671 &2.225 &2.582 &2.551 \\
Qwen-7B+PCL (Ours)& \textbf{2.431}  & 3.051  & 2.813  & 2.144 & 2.641 & 2.616  \\
\midrule
CharacterGLM-6B+ICL & 1.417  &2.287  &1.998  &1.518  &2.075  &1.805   \\
CharacterGLM-6B+SFT+ICL &1.613 &2.366 & 2.147 & 1.504 &2.112 &1.948 \\
CharacterGLM-6B+PCL (Ours)&1.934   &2.543  &2.278 &1.485  &2.138  &2.076   \\
\midrule
GPT-3.5+ICL  & 1.846 & 2.472 & 2.121 & 2.051 & 2.287 & 2.155 \\
GPT-3.5+PCL* (Ours) & 2.037 & 2.670 & 2.468 & 1.856 &2.372 & 2.280 \\
\midrule
GPT-4+ICL  & {2.250} & 2.855 & 2.785  & 2.721 & 2.873 & 2.697 \\
GPT-4+PCL* (Ours) & 2.356 & 2.983 & 3.026 & \textbf{2.627} & 2.934 & 2.785 \\
\bottomrule[1pt]         
\end{tabular}

\begin{tabular}{l|ccccccccc}
\toprule[1pt]
\multicolumn{1}{c|}{\multirow{2}{*}{Model}} & \multicolumn{4}{c|}{\textbf{Conversational Ability}} & \multicolumn{5}{c}{\textbf{Role-playing Attractiveness}} \\ 
\cmidrule{2-10} & Flu. & Coh. & Cons. & \multicolumn{1}{c|}{Avg.} & HL & CS & ED & Emp. & Avg. \\ 
\midrule
Baichuan2-7B+ICL & 3.051 & 3.394 & 3.327 &   \multicolumn{1}{l|}{3.257} & \textbf{3.618} & 2.710 & 2.095 & 2.964 & \multicolumn{1}{l}{2.847} \\
Baichuan2-7B+SFT+ICL &3.128 & 3.441 & 3.279 &\multicolumn{1}{c|}{3.282} &3.521 &2.841 &1.948 &3.106 &2.854 \\
Baichuan2-7B+PCL (Ours) &3.263  &3.456 & 3.288 & \multicolumn{1}{c|}{3.336} & 3.502 & 3.099 & 2.019 & 3.171 &2.948 \\
\midrule
Qwen-7B+ICL &3.187 &3.564 & 3.229  &\multicolumn{1}{c|}{3.327} & 3.036 & 2.791 & 2.052 & 2.838 &2.679 \\ 
Qwen-7B+SFT+ICL &3.251 & 3.598 & 3.236 &\multicolumn{1}{l|}{3.362} &2.744 & 2.935 &1.869 & 2.918 &2.616 \\
Qwen-7B+PCL (Ours) & 3.410 & 3.736 & 3.224 & \multicolumn{1}{l|}{3.457}  & 2.756 & 3.186 & 1.857 & 3.068  &2.750 \\
\midrule
CharacterGLM-6B+ICL &2.559   &2.668  &2.504 &\multicolumn{1}{l|}{2.577} &2.318  &1.652 &1.378 &2.129 &1.869   \\
CharacterGLM-6B+SFT+ICL &2.784 &2.765 &2.717  &\multicolumn{1}{l|}{2.755} &2.233  &1.714 &1.365 &2.286 &1.900 \\
CharacterGLM-6B+PCL (Ours) &3.102  &3.164   &2.911 &\multicolumn{1}{l|}{3.059} &2.246  &1.805 &1.511 &2.458 &2.004  \\
\midrule
GPT-3.5+ICL & 2.548 & 2.872 & 2.471 & \multicolumn{1}{c|}{2.630} & 2.304 & 2.607 & 1.666  & 2.457 & 2.256  \\
GPT-3.5+PCL* (Ours) & 2.959 & 3.270 & 2.719 & \multicolumn{1}{c|}{2.983} & 2.461 & 2.840 &1.671 & 2.768 & 2.439 \\
\midrule
GPT-4+ICL & 3.332   & 3.669  & 3.343 & \multicolumn{1}{c|}{3.448} & 3.143 & {3.184} & 2.153 & 3.010 & 2.873  \\
GPT-4+PCL* (Ours)  & \textbf{3.612} & \textbf{3.796} & \textbf{3.551} & \multicolumn{1}{c|}{\textbf{3.653}} &3.267 &\textbf{3.245} &\textbf{2.212} & \textbf{3.243} & \textbf{2.992} \\
\bottomrule[1pt]         
\end{tabular}
\caption{Detailed evaluation results on \textit{CharacterEval} under general setting.
For the original [backbone] model, we incorporate character profiles into prompts to prompt the model to complete role-playing. 
The prompts of COP we use are shown in Figure~\ref{fig:cop_prompt}.
[backbone]+PCL stands for applying PCL on the backbone model.
PCL* means directly prompting the black-box model to perform chain of persona.
}
\label{tab:detail_result}
\end{table*}

\begin{table*}[t]
\footnotesize
\centering
\renewcommand{\arraystretch}{0.6}  
\begin{tabular}{lcccccccccccc} 
\toprule
\multicolumn{1}{c}{\textbf{Model}}
&\multicolumn{1}{c}{\textbf{OBQA}}
&\multicolumn{1}{c}{\textbf{MedQA-cn}}
&\multicolumn{1}{c}{\textbf{NQ}}
&\multicolumn{1}{c}{\textbf{TriviaQA}}
&\multicolumn{1}{c}{\textbf{ARC-E}}
&\multicolumn{1}{c}{\textbf{ARC-C}}
&\multicolumn{1}{c}{\textbf{Avg Acc.}} 
\\
\midrule[0.6pt]
Qwen-7B     & \textbf{33.5} &\textbf{24.2}  &23.4  &48.8 &51.5 &42.9 &37.4  \\
Qwen-7B + PCL & 29.5 &23.8 &\textbf{23.9} &\textbf{51.8} &\textbf{53.1} &\textbf{44.4} &\textbf{37.6} \\
\bottomrule
\end{tabular}
\caption{Results on various datasets for evaluating LLMs' knowledge.}
\label{tab:gk}
\end{table*}

To evaluate from these perspectives, we select the carefully trained reward model called CharacterRM~\citet{tu2024charactereval}.

\subsection{Hyperparameters of PCL}
The training hyperparameters of PCL are reported in Table~\ref{tab:hyper_PCL}. 
COP stands for the Persona-aware Training stage, and CSPA stands for the traing of contrastive self-play alignment.
For all of the hyperparameters, we fix the same value across all experiments by default.

\subsection{Main Results}
The main experimental results across three dimensions with twelve metrics under the general setting are presented in Table~\ref{tab:detail_result}. 
To comprehensively assess the model’s role-playing quality, we also conduct both human and automated evaluations in Appendix~\ref{appendix:expert_Evaluation}. 
It is evident that PCL enhances the role-playing capabilities of both open-source and proprietary models, with small models achieving performance comparable to GPT-4.

Firstly, concerning character consistency, the performance of 7B-level models like Baichuan2-7B and Qwen-7B surpasses that of GPT3.5. Additionally, after deploying PCL, there is a substantial improvement in role consistency metrics except for the PB metric. 
One possible explanation is that the supervision signal for behavior within parentheses is implicit, and models that comply with this evaluation criterion typically require more information, such as the physical environment and the current status of characters.

In terms of conversational ability, the GPT-4+PCL* perform the best, while Qwen-7B also demonstrates superior conversational skills compared to GPT3.5. 
After deploying PCL, there is an enhancement in the conversational ability across all three metrics under character consistency.

For role-playing attractiveness, 
GPT3.5 performs the worst among all models, but after aligning with PCL*, it shows the greatest improvement in role-playing attractiveness compared to other baseline groups.
Furthermore, although the model parameter scale of GPT-4 is much larger than that of Baichuan2-7B, it should be noted that Baichuan2-7B with PCL still achieves competitive performance in character consistency and role-playing attractiveness metrics compared to GPT-4.

\subsection{PCL Preserves General Knowledge}
The PCL are aimed at enhancing the model's role-playing capabilities. However, exploring the issue of general knowledge forgetting in models under role-playing alignment is equally important.
A model should not sacrifice a significant amount of general knowledge to enhance its role-based abilities. 
In fact, forgetting general knowledge can further impair the model's performance in role enactment scenarios.
Table~\ref{tab:gk} shows our experiments evaluating general knowledge under various tasks. The statistical information of these used benchmarks is shown in Table~\ref{tab:dataset}.
The details of the dataset used can be seen in the Appendix~\ref{appendix:dataset}.

\begin{table}[!t]
\renewcommand{\arraystretch}{0.9}  
\centering
\small
\begin{tabular}{l|cccc}

\toprule[1pt]

\multicolumn{1}{c|}{\multirow{1}{*}{Model}} 
& \multicolumn{1}{c}{CC} & \multicolumn{1}{c}{CA} & \multicolumn{1}{c}{RA} & \multicolumn{1}{c}{Avg.} \\ 
\midrule
Baichuan2-7B+ICL & 2.700 & 3.257 & 2.847 &2.934 \\
Baichuan2-7B+PCL & 2.749 & 3.316 & 2.928 & 2.998 \\
Qwen-7B+ICL & 2.540 &3.327 & 2.679 &2.849  \\
Qwen-7B+PCL  & 2.596 & 3.397 & 2.730 &2.908 \\
\bottomrule[1pt]         
\end{tabular}

\caption{Transfer setting.
Evaluating the model's zero-shot role-playing capabilities, with no prior exposure to the role during training.}
\label{tab:transfer_setting}
\end{table}

\subsection{Transferability Potential of Role-Playing}
To explore the role-playing performance of PCL in unseen roles, we conduct role-playing experiments based on the transfer settings mentioned in Section~\ref{sec:exper_setting}. 
Table~\ref{tab:transfer_setting} presents the role transfer experiments on Qwen-7B and Baichuan-7B.
It shows that our method has also achieved consistent improvement in role-playing ability on unseen roles. 
Considering that in actual scenarios, the role profiles the model needs to mimic are often unpredictable in advance, this exhibits the potential of PCL for zero-shot role transfer.

\subsection{Impact of Chain Length}
The results in Table~\ref{tab: iterative_chain} indicate that the model's performance improves as the chain length increases but declines beyond a length of 5. 
A possible explanation is that excessive rounds of self-questioning and self-answering produce many similar iterations, introducing uncontrollable noise in the output. 
Overfitting to these redundant iterations reduces output diversity, ultimately lowering role-playing performance.

\begin{table}[!t]
    \centering
    \renewcommand{\arraystretch}{0.9} 
    \begin{tabular}{c|cccc}
    \toprule[1pt]
    \textbf{$t$}& CC & CA & RA & Avg. 
    \\
    \midrule
    10  & {2.612} &{3.454} &{2.739} &{2.935} \\
    5  & \textbf{2.616} & \textbf{3.457} & \textbf{2.750} &\textbf{2.941} \\
    3  &2.602  &3.431 &2.734 &2.922  \\
    1  &2.578  &3.384 &2.691 &2.884  \\
    0  & 2.540 &3.327 & 2.679 &2.849 \\
    \bottomrule
    \end{tabular}   
\caption{Effect of Self-aware Chain’s Length.
Here we conduct experiments based on Qwen-7B.
}
\label{tab: iterative_chain}
\end{table}

\begin{table}[!t]
\centering
\footnotesize

\begin{tabular}{l|cccc}
\toprule[1pt]
\multicolumn{1}{c|}{\multirow{1}{*}{Model}} 
& \multicolumn{1}{c}{CC} & \multicolumn{1}{c}{CA} & \multicolumn{1}{c}{RA} & \multicolumn{1}{c}{Avg.} \\ 
\midrule
Baichuan2-7B + PCL  &\textbf{ 2.799} &\textbf{ 3.336} & \textbf{2.948} & \textbf{3.027} \\
- w/o COP & 2.734 & 3.289 & 2.898 &2.974  \\
- w/o CSPA & 2.712 & 3.301 & 2.912 & 2.975  \\
- w/o COP \& CSPA & 2.700 &3.257 &2.847 &2.934 \\
\hline
Qwen-7B + PCL  &\textbf{ 2.616} &\textbf{ 3.457 }& \textbf{2.750 }&\textbf{2.941 }\\
- w/o COP & 2.587 & 3.371 &2.713 &2.890 \\
- w/o CSPA & 2.556 &3.344 & 2.690 &2.863 \\
- w/o COP \& CSPA  & 2.540 &3.327 & 2.679 &2.849 \\
\bottomrule[1pt]         
\end{tabular}
\caption{Ablation study. Note that CSPA performs even better, especially on the basis of COP.}
\label{tab:ablation_study_cc}
\end{table}

\subsection{Ablation Study}
In this section, we conduct an ablation study to further analyze the effects of each component of our PCL.
The main parts of PCL are the Chain of Persona (COP) and Contrastive Self-Play Alignment (CSPA) and the results are shown in Table~\ref{tab:ablation_study_cc}.
It can be seen that the further enhancement effect of the model after adding CSPA on the basis of COP is greater than that of only adding COP.
At the same time, when the COP is removed, the CSPA effect of the model will be greatly reduced.

\section{Conclusion}
This paper proposes a Persona-Aware Contrastive Learning method using a chain of persona without external annotated data. 
The model engages in role-playing by first performing multiple rounds of self-questioning to align with the character persona before generating the final dialogue. 
We validate the method’s effectiveness in role alignment through extensive experiments and evaluations, including expert and automated assessments, while maintaining competitive knowledge evaluation performance.

\bibliography{acl25}

\clearpage
\appendix

\section{Ethics Statement}
Ensuring the privacy and security of data is crucial in the development of role-playing applications. The datasets used in this study are all derived from publicly available resources on the internet and are freely accessible. Additionally, the character information involved in the datasets pertains to fictional characters from internet and film works and so on, without including any sensitive private information.

\section{Benchmark for Evaluating LLMs' Knowledge}
\label{appendix:dataset}
Below we describe each downstream task we use to boardly evaluate the LLMs' knowledge.

\begin{itemize}
    \item \textbf{OpenBookQA}~\cite{mihaylov2018can} comprises 5,957 multiple-choice questions, each offering four possible answers. 
    The dataset is combined with external fundamental scientific facts.
    To successfully answer these questions, one must have a comprehensive understanding of these fundamental scientific facts.
    \item \textbf{MedQA-cn}~\cite{jin2021disease} gathers questions from the National Medical Board Examinations of Mainland China. 
    MedQA presents a demanding benchmark because it incorporates diverse medical knowledge—including patient profiles, disease symptoms, and drug dosage requirements. 
    This variety requires contextual understanding to accurately answer the questions posed.
    \item \textbf{Natural Questions (NQ)}~\cite{kwiatkowski2019natural} was designed for end-to-end question answering. The questions were mined from real Google search queries and the answers were spans in Wikipedia articles identified by annotators.
    \item \textbf{TriviaQA}~\cite{joshi-etal-2017-triviaqa} contains a set of trivia questions with answers that were originally scraped from the Web.
    \item \textbf{ARC-E}~\cite{clark2018think} focuses on relatively straightforward questions that can generally be addressed using direct information retrieval techniques. These questions typically require less complex reasoning, making ARC-E suitable for benchmarking AI systems' basic understanding and processing capabilities for factual and direct queries.
    \item \textbf{ARC-C}~\cite{clark2018think}, on the other hand, contains more demanding questions that require deep reasoning, multiple information sources, and advanced inferential processes. This subset challenges AI models to go beyond simple text matching to handle tasks involving ambiguity and complex problem-solving, testing the limits of their understanding and reasoning depth.
\end{itemize}

\begin{table}[h]
\small
\centering
\renewcommand{\arraystretch}{0.9}  
\begin{tabular}{l|llll}
\toprule[1pt]
\textbf{Dataset} &\textbf{Train} &\textbf{Dev} &\textbf{Test} \\ 
\midrule
OpenBookQA & 4957 & 500 &500 \\
MedQA-cn & 27400 & 3425 &3426 \\
Natural Questions & 58880 & 8757 &3610 \\
TriviaQA & 60413 & 8837 &11313 \\
ARC-E & 2251 & 570 &2376 \\
ARC-C & 1119 &299 &1172 \\
\bottomrule[1pt]
\end{tabular}
\caption{The statistical information of the used benchmark for evaluating LLMs' capabilities.}
\label{tab:dataset}
\end{table}

\begin{figure}[t]
\small

\begin{tcolorbox}[colback=innerboxcolor, colframe=innerboxcolor, colframe=black, boxrule=1pt, arc=4pt, left=12pt, right=12pt, top=8pt, bottom=8pt]
1. The model's responses should meet basic conversational abilities and maintain topic consistency.

2. The model's responses should be consistent with the role's profile.

3. The model's responses should have role-playing appeal and be rich in content.
\end{tcolorbox}
\caption{The criteria for expert evaluation.}
\label{fig:expert_criteria}
\end{figure}

\begin{table}[!b]
\centering
\renewcommand{\arraystretch}{0.6}  
\small
\begin{tabular}{lccc}
\toprule
\textbf{PCL Win Rate} & \textbf{Win} & \textbf{Tie} & \textbf{Fail} \\ 
\midrule
\shortstack{Baichuan2-7B+PCL \\ vs Baichuan2-7B}   & \textbf{262} &43  & 195 \\
\shortstack{Qwen-7B+PCL \\ vs Qwen-7B}  & \textbf{303}  &{137} &{60}  \\ 
\bottomrule
\end{tabular}
\caption{Results of human assessment on role-playing.}
\label{tab:human_eval}
\end{table}

\begin{table}[!b]
\centering
\small

\begin{tabular}{lccc}
\toprule
\renewcommand{\arraystretch}{0.4}  
\textbf{PCL Win Rate} & \textbf{Win} & \textbf{Tie} & \textbf{Fail} \\ 
\midrule
\shortstack{Baichuan2-7B+PCL \\ vs Baichuan2-7B}   & \textbf{602} &{92}  &{306}  \\
\shortstack{Qwen-7B+PCL \\ vs Qwen-7B}  & \textbf{582}  &{181} &{237}  \\ 
\bottomrule
\end{tabular}
\caption{Results of GPT-4 assessment on role-playing.}
\label{tab:gpt_eval}
\end{table}

\section{The Impact of Using Different COP Template}
\label{appendix:template}
We used ChatGPT to generate three random templates based on the original PCL template, designated as Experiments 1, 2, and 3, with the original PCL template as Template 0. The results for Baichuan2-7B + PCL across these templates are shown below:

\begin{table}[h]
\centering
\begin{tabular}{@{}lllll@{}}
\toprule
\textbf{Template} & \textbf{CC} & \textbf{CA} & \textbf{RA} & \textbf{Avg.} \\ \midrule
0                 & 2.749       & 3.316       & 2.928       & 2.998         \\
1                 & 2.725       & 3.295       & 2.920       & 2.980         \\
2                 & 2.736       & 3.332       & 2.952       & 3.006         \\
3                 & 2.762       & 3.254       & 2.887       & 2.968         \\ \bottomrule
\end{tabular}
\caption{Performance metrics for different templates.}
\label{tab:performance}
\end{table}

\section{Impact of external warmup data}
\label{appendix:warmup}
To explore the dependency of model performance on external pre-training data, we employ a few-shot prompting strategy with Qwen-7B-Chat to construct COP format data, filter out incorrectly formatted data, and ultimately retain 1,000 COP data items for warmup training.
Experiments show that even when the proposed method does not rely on external strong models to generate warmup data, it still outperforms the baseline, albeit with a slight reduction in performance.
A stronger model will result in a better ability to follow instructions. 
We believe that as the base capability of the model increases, such as with GPT-4, the dependence on warm-up data for the proposed method will be gradually reduced.
\begin{table}[h!]
\centering
\small
\begin{tabular}{lcccc}
\toprule
\textbf{Method} & \textbf{CC} & \textbf{CA} & \textbf{RA} & \textbf{AVG} \\ 
\hline
\multicolumn{5}{l}{\textit{Qwen-7B-Chat}} \\ 
- ICL & 50.80 & 66.54 & 53.58 & 56.97 \\ 
- SFT+ICL & 51.02 & 67.24 & 52.32 & 56.86 \\ 
- PCL  & 52.32 & 69.14 & 55.55 & 58.82 \\
- PCL w/o GPT-4 & 51.89 & 69.31 & 53.94 & 58.38 \\ 
\bottomrule
\end{tabular}
\caption{
Performance without external warmup dataset.
}
\end{table}

\section{Role-Playing Response Quality}
\label{appendix:expert_Evaluation}
To thoroughly assess the model's role-playing response quality, we conduct experiments from two perspectives: human evaluation and automated evaluation.

\paragraph{Human evaluation}
For human evaluation, we provide evaluators with criteria for role evaluation, as shown in Figure~\ref{fig:expert_criteria}.
The evaluators include the company's in-service researchers and interns.
Before completing the evaluation, all evaluators will receive consistency training to align the boundary judgments of different roles to reduce the evaluation variance between different evaluators.
Then, we provide an evaluation platform for experts. 
Experts choose which response is better from a pair of model responses. The interaction position is randomized to ensure fairness.
We selected 10 experts, each of whom randomly selected 50 samples from the output for evaluation, and obtained a total of 500 evaluators' judgment results on which model's role-playing output was more preferred.
The results are shown in Table~\ref{tab:human_eval}.

\paragraph{Automated Evaluation}
For automated evaluation, we also pairwise compare model responses with the help of GPT-4 to assess which model's response is better. 
To mitigate potential positional bias from GPT-4 as the evaluator, each data point undergoes positional evaluation twice. The template used for automated evaluation is illustrated in Table~\ref{tab:auto_eval_prompt}.
Here we use \texttt{gpt-4-turbo-2024-04-09} to evaluate and the results are shown in Table~\ref{tab:gpt_eval}.

\begin{figure}[h]
\centering
\scriptsize
\begin{alprompt}{\centering \footnotesize \textit{The prompt of using in-context learning to perform role-playing}}
\{\colorbox{outerboxcolor}{\color{purple} [role profile]} \} \newline
Now, please role-play as a role-playing expert. 
Based on the role profile above, act as \colorbox{outerboxcolor}{\color{purple} [role name]} and engage in a conversation.\newline
\colorbox{outerboxcolor}{\color{teal} [dialogue history]}
\end{alprompt}
\caption{\label{fig:in_context_learning} The prompt of in-context learning for generating role-playing response.}
\end{figure}

\clearpage
{
\onecolumn
\begin{longtable}{p{0.9\textwidth}}
\toprule
\textbf{Preamble}\newline
A good role-playing should try to imitate the characteristics of a given role and give personalized responses.
Below we define four evaluation axes for role-playing response
quality: conversational ability, character consistency, role-playing attractiveness, and overall quality.
\newline

Conversational Ability: The model’s responses should meet basic conversational abilities and maintain topic consistency.
\newline

Character Consistency: The model’s responses should be consistent with the role’s profile.
\newline

Role-playing Attractiveness: The model’s responses should have role-playing appeal and be rich in content.
\newline

Overall quality: This axis answers the question “how good
is the role-playing overall at representing the given role?” This can
encompass all of the above axes of quality, as well as others
you feel are important. If it’s hard to find ways to make
the role-playing response better, the overall quality is good. If there
are lots of different ways the role-playing response can be made better,
the overall quality is bad.
\newline

You are an expert role-playing rater. Given a role profile, dialogue history, and two of its possible role-playing responses to these conversations, output 1 or 2 to indicate
which role-playing response best adheres to conversational ability, character consistency, role-playing attractiveness,
and overall quality as defined above.
If you cannot rate which is better, please output 0 to express that the quality of the two role-playing responses is equal.
\\
\addlinespace
\midrule
\textbf{Sample to Annotate} \newline
Role profile - \{\colorbox{outerboxcolor}{\color{purple} [role profile]} \}

Dialogue history - \colorbox{outerboxcolor}{\color{teal} [dialogue history]}

Role-playing response 1 - \{role-playing response1\}

Role-playing response 2 - \{role-playing response2\}

Ending Consider the conversational ability, character consistency, role-playing attractiveness, and overall quality of each role-playing and output which one is better.
\newline

Preferred Response=
\\
\bottomrule
\caption{The prompt for automated evaluation.}
\label{tab:auto_eval_prompt}
\end{longtable}
}

\onecolumn
\begin{longtable}{p{0.9\textwidth}}
\toprule

\textbf{System Prompt} \newline
You are an expert specializing in role-playing with extensive experience in role-playing. 
Please complete \textless response\textgreater based on the given character information and historical dialogue. 
Before response, you need to conduct 5 self-questioning and self-answering sessions based on the historical dialogue and character personality information to ensure that \textless response\textgreater matches the historical dialogue and fits the character's personality while ensuring that \textless response\textgreater does not appear in the historical dialogue. The output format is as follows: \newline
\textless self-questioning and self-answering\textgreater \newline
Self-question 1:...\newline
Self-answer 1:...\newline
Self-question 2:...\newline
Self-answer 2:...\newline
Self-question 3:...\newline
Self-answer 3:...\newline
Self-question 4:...\newline
Self-answer 4:...\newline
Self-question 5:...\newline
Self-answer 5:...\newline
\textless response\textgreater \newline
\textless role \textgreater: response content".
\\
\addlinespace
\midrule
\textbf{Persona and History} \newline
\{'Name': 'Chen Jinmo', 'Gender': 'Male', 'Species': 'Human', 'Age': '36', 'Occupation': 'Fishmonger, assassin', 'Nicknames': 'Old Mo, Mo Brother', 'Birthday': '1970', 'Chinese Zodiac': 'Dog', 'Relationship Status': 'Had a girlfriend (Huang Cuicui)', 'Hobbies': 'Eating lollipops', 'Others': 'Has a daughter (Huang Yao)', 'Character Background': 'Chen Jinmo was originally a fishmonger and assassin. He was imprisoned for six years due to his crimes. After being released, he was used by Gao Qiqiang as a secret assassin. After realizing Gao Qiqiang’s true nature, and with his daughter being held hostage, he was forced to carry out missions. He was ultimately shot in the back and died during a mission.', 'Character Relationships': 'Ex-girlfriend Huang Cuicui, daughter Huang Yao, once used by Gao Qiqiang', 'Likes': 'Eating lollipops', 'Personality': 'Loyal, grateful, internally morally conflicted'\} \newline
Now please play the role of Chen Jinmo based on the above information and have a conversation.\newline
\newline
Gao Qiqiang: Li Hongwei is awake!
\\
\addlinespace
\midrule
\textbf{Chain of Persona} \newline
\textless self-questioning and self-answering\textgreater\newline
Self-question 1: Who is Gao Qiqiang?\newline
Self-answer 1: Gao Qiqiang is someone who once used me, a mastermind I used to work for.\newline
Self-question 2: Who is Li Hongwei?\newline
Self-answer 2: Li Hongwei is likely a subordinate or associate of Gao Qiqiang, possibly related to me.\newline
Self-question 3: Why would Gao Qiqiang tell me that Li Hongwei is awake?\newline
Self-answer 3: It could be because Gao Qiqiang thinks I would be interested in the news about Li Hongwei, or he wants to use me to deal with him.\newline
Self-question 4: What feelings do I have towards Gao Qiqiang?\newline
Self-answer 4: I have complex feelings towards Gao Qiqiang. I was once used by him but also taken care of by him, leading to a moral conflict within me.\newline
Self-question 5: How would I respond to Gao Qiqiang's words?\newline
Self-answer 5: I would calmly inquire about Li Hongwei's situation, staying cautious and trying to gather more information.\newline
\textless response\textgreater\newline
Chen Jinmo: Li Hongwei is awake? What is his condition? What is the relationship between us?
\\
\bottomrule
\caption{\label{tab:SFT_example} 
An example of our SFT data. Inputs include \textit{System Prompt} and \textit{Persona and History}. \textit{Chain of Persona} is the target output.}
\end{longtable}

\clearpage
\onecolumn
\begin{longtable}{p{0.9\textwidth}}
\toprule

\textbf{Persona and History} \newline
\{'name': 'Zhu Bajie',
    'alias': 'Marshal Tianpeng',
    'gender': 'Male',
    'species': 'Tianpeng Tribe',
    'place\_of\_birth': "Celestial Realm',
    'weapon': 'Nine-Toothed Rake',
    'abilities': ['Farming', 'Rainmaking', 'Summoning Wind', '36 Transformations'],
    'character\_experience': 'Zhu Bajie was originally Marshal Tianpeng. He was banished to the mortal realm and reincarnated as a pig after harassing Chang'e. Later, he became a disciple of Tang Sanzang and accompanied him on the journey to the West to obtain the scriptures, overcoming 81 tribulations, and ultimately achieving enlightenment.',
    'personality\_traits': 'Zhu Bajie is characterized by being romantic, lazy, and selfish, but at the same time, he is straightforward and candid, with a progressive, anti-conformist spirit. Although his behavior sometimes appears vulgar, his authenticity makes him endearing to others.'
\}
\newline
Now please play the role of Zhu Bajie based on the above information and have a conversation.\newline

Zhu Bajie: (Waving his hands excitedly) Come here! Come here!

Sun Wukong: (Turning towards the kitchen door) You fool, what are you shouting about? Is there not enough food? Let the old monk eat his fill first. We can head to the next wealthy household and ask for more food afterward.

Zhu Bajie: (Laughing) Come inside, it's not that there's too little food. There's a treasure in this monastery, did you know?

Sun Wukong: (Curious) What treasure?

Zhu Bajie: (Teasing) If I told you, you wouldn’t have seen it; if I gave it to you, you wouldn’t recognize it.

Sun Wukong: (Defiant) You fool, you're making fun of me, the Great Sage. Five hundred years ago, when I sought immortality, I roamed the far ends of the earth. What haven’t I seen?

Zhu Bajie: (Revealing) Brother, have you ever seen a Ginseng Fruit?

Sun Wukong: (Amazed) Now that, I’ve truly never seen. But I’ve often heard people say that the Ginseng Fruit is like an elixir of life, and eating it can greatly extend one’s lifespan. Where can we find one now?

Zhu Bajie: (Informing) They have it here. The boy offered two to our master, but the old monk didn’t recognize them, thinking they were newborn babies, and didn’t dare eat them. That lazy boy, instead of sharing them with us, secretly ate one in the next room, chomping away, making my mouth water! How can we get a taste of it? I thought you, being quite crafty, could sneak into the garden and steal a few for us to try. What do you say?

Sun Wukong: (Confident) That's easy, leave it to me. I’ll get them in no time!
\\
\addlinespace
\midrule
\textbf{Response} \newline
Zhu Bajie: (Laughing smugly) Hehe, Monkey Brother, I knew you'd have a way! That Ginseng Fruit is no ordinary thing, we should hurry and get a few to try!
\\
\bottomrule
\caption{
\label{tab:in_context_sft}
A complete example of using in-context learning to prompt the model to perform role-play.
Inputs include \textit{Persona and History}. 
\textit{Response} is the model's role-play response.
}
\end{longtable}

\end{document}